\documentclass{cap2026}

\usepackage{amsmath, amssymb}
\usepackage{booktabs}
\usepackage{graphicx}
\usepackage{algorithm}
\usepackage{algpseudocode}
\usepackage{xcolor}
\usepackage{float}
\usepackage{rotating}

\title{Opportunistic Target Selection: Early Directional Commitment for Query-Efficient Black-Box Adversarial Attacks}

\author[1]{Florent Tariolle}
\author[1,2]{Florian Yger}
\affil[1]{INSA Rouen Normandy}
\affil[2]{LITIS}
\date{\vspace{-2.5em}}

\begin{document}
\nolinenumbers

\maketitle

\begin{abstract}
Black-box adversarial attacks that minimize only the ground-truth confidence suffer from \textit{class drift}: perturbations wander through the feature space without committing to a specific adversarial class, wasting queries on diffuse, undirected progress. We introduce \textbf{Opportunistic Target Selection (OTS)}, a lightweight wrapper that switches an untargeted attack to a targeted objective early in its trajectory, locking onto whichever non-true class currently leads. OTS requires no architectural modification to the underlying attack, no gradient access, and no \textit{a priori} target-class knowledge.

We validate OTS on three score-based attacks (SimBA, Square Attack with cross-entropy loss, and Bandits) across five standard ImageNet classifiers (4,500 runs). On random-search attacks, OTS closely tracks oracle performance, with gains up to +27~pp in success rate and 43\% relative reduction in censored-mean iterations on ResNet-50. On gradient-estimation attacks (Bandits) and attacks with margin loss, OTS is redundant, a negative result that reinforces our interpretation of OTS as a margin-loss surrogate. On adversarially-trained models, a bimodal difficulty distribution eliminates the regime where targeting helps. We release our code at \url{https://github.com/Tariolle/opportunistic-target-selection}.

\medskip
\end{abstract}

\section{Introduction}\label{sec:introduction}

Standard untargeted black-box attacks minimize the model's confidence in the ground-truth class. This strategy, whether implemented as probability minimization in SimBA \citep{guo2019simple} or cross-entropy loss in Square Attack \citep{andriushchenko2020square}, treats all non-true classes as interchangeable. The freed probability mass spreads across remaining classes without directional commitment; the perturbation executes a random walk through the latent space, crossing class basins opportunistically rather than heading toward a specific decision boundary.

This \textit{class drift} directly impacts query efficiency. Each query spent exploring a class basin that will ultimately be abandoned is wasted. The harder the model is to attack, the more basins the perturbation traverses before settling, and the more pronounced the waste becomes.

Targeted attacks eliminate drift by construction but require specifying a target class \textit{a priori}, which is generally unavailable in a black-box setting. Margin-based losses \citep{carlini2017towards} offer a partial solution by implicitly tracking the nearest decision boundary, but not all attacks support them. SimBA operates on raw probabilities; other attacks use cross-entropy for compatibility. For these methods, drift remains the main efficiency bottleneck.

\textbf{Opportunistic Target Selection} bridges this gap with a two-phase strategy:
\begin{enumerate}
    \item \textbf{Exploration phase.} The attack runs in untargeted mode for a short prefix of $T$ iterations.
    \item \textbf{Exploitation phase.} The attack switches to a targeted objective against the leading non-true class, and runs until misclassification or budget exhaustion.
\end{enumerate}

The switch timing $T$ is not sensitive: ablations show that any $T \in \{5, \ldots, 500\}$ yields no statistically significant difference (Wilcoxon signed-rank tests, Section~\ref{sec:ablations}), because class rankings stabilize within the first few iterations for drift-prone attacks. This paper makes three contributions:

\begin{enumerate}
    \item \textbf{OTS as a margin-loss surrogate.} By locking onto the leading non-true class early in the trajectory, OTS approximates the competitor term of margin loss for attacks that lack implicit target tracking. This explains both where OTS helps (probability-minimization and CE losses) and where it is redundant (margin loss, gradient estimation), revealing that the information needed to select an effective target is already latent in the attack's trajectory after just a few iterations.
    \item \textbf{Empirical validation} across five ImageNet classifiers and three attack methods (4,500 runs), showing that OTS provides near-oracle efficiency. Target selection is critical (random targeting is catastrophic), and a short exploration phase provides cheap insurance over the clean-image argmax.
    \item \textbf{A characterization of targeting neutrality} on adversarially-trained models, where a bimodal difficulty distribution eliminates the medium-difficulty regime where directional commitment provides value.
\end{enumerate}

\section{Related Work}\label{sec:related-work}

Score-based black-box attacks access the full output probability vector but not gradients. SimBA \citep{guo2019simple} iterates over an orthonormal basis, accepting perturbations that reduce the true-class probability. Square Attack \citep{andriushchenko2020square} uses random square patches with a shrinking schedule; its default margin loss implicitly tracks the nearest decision boundary, but switching to cross-entropy loss removes this guidance and introduces drift. Gradient-estimation attacks such as Bandits \citep{ilyas2019prior} incorporate data-dependent priors, providing built-in directionality analogous to margin loss. Decision-based attacks (SurFree \citep{maho2021surfree}, CGBA \citep{gesny2024gradient}) operate with top-1 labels only; we adopt the angular convergence framework of \citet{gesny2024gradient} in Section~\ref{sec:theta-convergence}. To our knowledge, no prior work has studied online target selection during a score-based attack's trajectory; existing targeted attacks assume the target class is specified before the attack begins. Adversarially-trained models \citep{madry2018towards,salman2020adversarially}, standardized via RobustBench \citep{croce2021robustbench}, present a qualitatively different challenge; Section~\ref{sec:robust} examines OTS in this regime.

\section{Method}\label{sec:method}

\subsection{Notation}\label{sec:notation}

Let $f: \mathcal{X} \to \mathbb{R}^K$ denote a classifier mapping inputs to logits over $K$ classes, and $P(k|x) = \text{softmax}(f(x))_k$ the predicted probability of class $k$. Given a correctly-classified input $x$ with true label $y$, the attacker seeks an adversarial example $x' = x + \delta$ such that $\arg\max_k P(k|x') \neq y$ and $\|\delta\|_\infty \leq \epsilon$.

\subsection{Untargeted Loss and the Drift Problem}\label{sec:drift}

Standard untargeted attacks minimize a loss that depends only on the true class:
\begin{equation}
\mathcal{L}_{\text{untarg.}}(x', y) = P(y | x') \;\text{(SimBA)} \quad \text{or} \quad \log P(y | x') \;\text{(CE)}
\end{equation}
These objectives decrease $P(y|x')$ without specifying where the freed probability mass should concentrate. We call this \textit{class drift}. \textbf{Margin loss} avoids drift by construction:
\begin{equation}
\mathcal{L}_{\text{margin}}(x', y) = f_y(x') - \max_{k \neq y} f_k(x')
\end{equation}
where $f_k$ denotes the logit for class $k$ (as opposed to the probability-space objectives above) and $\max_{k \neq y} f_k(x')$ dynamically identifies the nearest competitor. Square Attack's default loss is of this form, explaining its strong untargeted performance.

A targeted attack toward class $t$ optimizes $\mathcal{L}_{\text{targeted}}(x', t) = -P(t | x')$ (SimBA) or $\log P(t | x')$ (CE). Targeting eliminates drift but requires knowing $t$ in advance. An \textit{oracle} target, defined as the class the unconstrained untargeted attack eventually reaches, provides a trajectory-specific performance ceiling that no real attacker can achieve.

\subsection{Opportunistic Target Selection Algorithm}\label{sec:algorithm}

OTS discovers the target class online by observing which adversarial class the perturbation is naturally drifting toward.

\textbf{Fixed-iteration switch (recommended).} The attack runs in untargeted mode for $T$ iterations, then switches to a targeted objective against $t = \arg\max_{k \neq y} P(k|x')$. Ablations (Section~\ref{sec:ablations}) show that $T \in \{5, \ldots, 500\}$ all yield indistinguishable results, so any small $T$ suffices.

\textbf{Rank-stability variant.} A stability counter tracks the leading non-true class after each accepted perturbation step, switching only after the same class leads for $S$ consecutive accepted steps (Algorithm~\ref{alg:ot}). In practice, this provides no measurable advantage over the fixed switch (Section~\ref{sec:ablations}). We retain it for completeness.

\begin{algorithm}[t]
\caption{Opportunistic Target Selection (Rank-Stability Variant)}
\label{alg:ot}
\begin{algorithmic}[1]
\Require image $x$, true label $y$, attack $A$, threshold $S$
\Ensure adversarial example $x'$
\State $x' \leftarrow x$;\; $t \leftarrow \bot$;\; $c \leftarrow \bot$;\; $s \leftarrow 0$
\While{$\arg\max_k P(k|x') = y$ \textbf{and} budget remaining}
    \If{$t = \bot$}
        \State $x' \leftarrow A.\text{step}(x', y)$
        \State $\ell \leftarrow \arg\max_{k \neq y} P(k|x')$
        \If{$\ell = c$} $s \leftarrow s + 1$ \Else{} $c \leftarrow \ell$;\; $s \leftarrow 1$ \EndIf
        \If{$s \geq S$} $t \leftarrow c$ \EndIf
    \Else
        \State $x' \leftarrow A.\text{step}(x', t)$
    \EndIf
\EndWhile
\State \Return $x'$
\end{algorithmic}
\end{algorithm}

Once a target is locked, the attack commits for the remainder of the budget (\textbf{irreversible lock}). Releasing the lock would re-introduce the exploration overhead that OTS is designed to eliminate.

\subsection{Attack Integration}\label{sec:integration}

\textbf{SimBA.} We use the DCT basis with $8\times8$ blocks, $\epsilon = 8/255$ in $L_\infty$, and a 10,000-iteration budget. In untargeted mode, SimBA reduces $P(y|x')$; upon lock-in, the criterion switches to increasing $P(t|x')$. The perturbation mechanism is unchanged.

\textbf{Square Attack (CE loss).} We deliberately run Square Attack with cross-entropy loss rather than its default margin loss to isolate OTS's contribution: margin loss already provides implicit target tracking (Section~\ref{sec:drift}), which would confound OTS's effect. Budget is $N = 10{,}000$ iterations; the patch schedule depends on $N$, so all experiments use a fixed $N$ for comparability.

\textbf{Bandits.} Bandits \citep{ilyas2019prior} estimates gradients with data-dependent priors, providing built-in directionality analogous to margin loss. We use the stability variant ($S = 15$), 5,000-iteration budget, $\epsilon = 8/255$.

\section{Experimental Setup}\label{sec:setup}

\hspace{\parindent}\textbf{Models.} We evaluate on five standard ImageNet classifiers from torchvision (AlexNet, ResNet-18, VGG-16, ResNet-50, ViT-B/16 \citep{dosovitskiy2021image}); all models are wrapped so that attacks operate in $[0, 1]$ pixel space. ViT-B/16 tests whether OTS's benefit depends on convolutional inductive biases. We also evaluate two adversarially-trained models from RobustBench \citep{salman2020adversarially,croce2021robustbench}: Salman2020Do\_R18 and Salman2020Do\_R50.

\textbf{Protocol.} Each (model, attack, image) triplet is evaluated in three modes: \textit{untargeted} (standard attack), \textit{targeted oracle} (target class chosen \textit{a posteriori} from the untargeted result, providing a trajectory-specific ceiling), and \textit{opportunistic} (OTS). The oracle is determined by running a probe untargeted attack with the same random seed; it is not globally optimal (a geometry-based target might yield shorter paths) but provides a reproducible reference for the perturbation path the attack would have followed.

\textbf{Configuration.} Perturbation budget $\epsilon = 8/255$ ($L_\infty$). Iteration budget: 10,000 (main benchmark) and 15,000 (extended benchmark). Each iteration corresponds to one model query for Square Attack and Bandits. SimBA uses two queries per iteration (one per sign direction), averaging 1.4--1.5 effective queries per iteration after early rejection. All budgets and iteration counts reported in this paper refer to iterations, not raw queries. Stability thresholds: $S = 10$ (SimBA), $S = 8$ (Square Attack) on standard models; $S = 10$ for both on robust models.

\textbf{Images.} 100 images randomly sampled from ILSVRC2012 validation (seed 42). Each combination uses a single attack random seed (seed 0), yielding 4,500 runs on standard models. A separate multi-seed validation (5 seeds) confirms seed independence (Figure~\ref{fig:multiseed}).

\textbf{Metrics.} Primary: \textit{success rate} at equal query budgets. We also report \textit{success rate vs.\ query budget} (CDF curves) following \citet{ughi2021empirical}. Secondary: \textit{censored mean iterations} (failed runs are assigned the budget ceiling as cost, so the metric penalizes both slow convergence and outright failure) and \textit{paired mean iterations} (restricted to images where both modes succeed, isolating pure efficiency). Censoring biases \emph{against} OTS, making savings estimates conservative.

\section{Results on Standard Networks}\label{sec:results}

\subsection{Success Rates}\label{sec:success-rates}

\begin{table}[H]
\centering
\caption{Success rates (\%) by model and attack mode (100 images per model, 10K budget).}
\label{tab:success-rates}
\small
\begin{tabular}{@{}l ccc ccc ccc@{}}
\toprule
& \multicolumn{3}{c}{SimBA} & \multicolumn{3}{c}{Square Attack (CE)} & \multicolumn{3}{c}{Bandits} \\
\cmidrule(lr){2-4} \cmidrule(lr){5-7} \cmidrule(lr){8-10}
Model & Unt. & OTS & Orac. & Unt. & OTS & Orac. & Unt. & OTS & Orac. \\
\midrule
AlexNet    & 77 & 78 & 78 & 100 & 100 & 100 & 47 & 46 & 47 \\
ResNet-18  & 85 & 90 & 93 & 100 & 100 & 100 & 37 & 35 & 33 \\
VGG-16     & 84 & 85 & 89 & 100 & 100 & 100 & 49 & 41 & 43 \\
ResNet-50  & 43 & 70 & 70 &  81 &  94 &  95 & 25 & 24 & 25 \\
ViT-B/16   & 46 & 52 & 51 & 100 & 100 & 100 & 28 & 24 & 25 \\
\bottomrule
\end{tabular}
\end{table}

Table~\ref{tab:success-rates} reports success rates across 4,500 runs (pooled: SimBA 67/75/76\%, Square Attack 96/99/99\%, Bandits 37/34/35\% for Unt./OTS/Orac.). For SimBA and Square Attack (CE), OTS closely tracks oracle performance. The largest gains appear on ResNet-50: +27~pp for SimBA (43\% $\to$ 70\%) and +13~pp for Square Attack (81\% $\to$ 94\%). Easier models (AlexNet, VGG-16, ResNet-18) show floor effects with near-perfect rates regardless of mode. Bandits shows no benefit from OTS and is occasionally harmed (VGG-16: 49\% $\to$ 41\%), confirming that gradient estimation already provides the directionality OTS adds.

\begin{figure}[ht]
\centering
\includegraphics[width=0.85\textwidth]{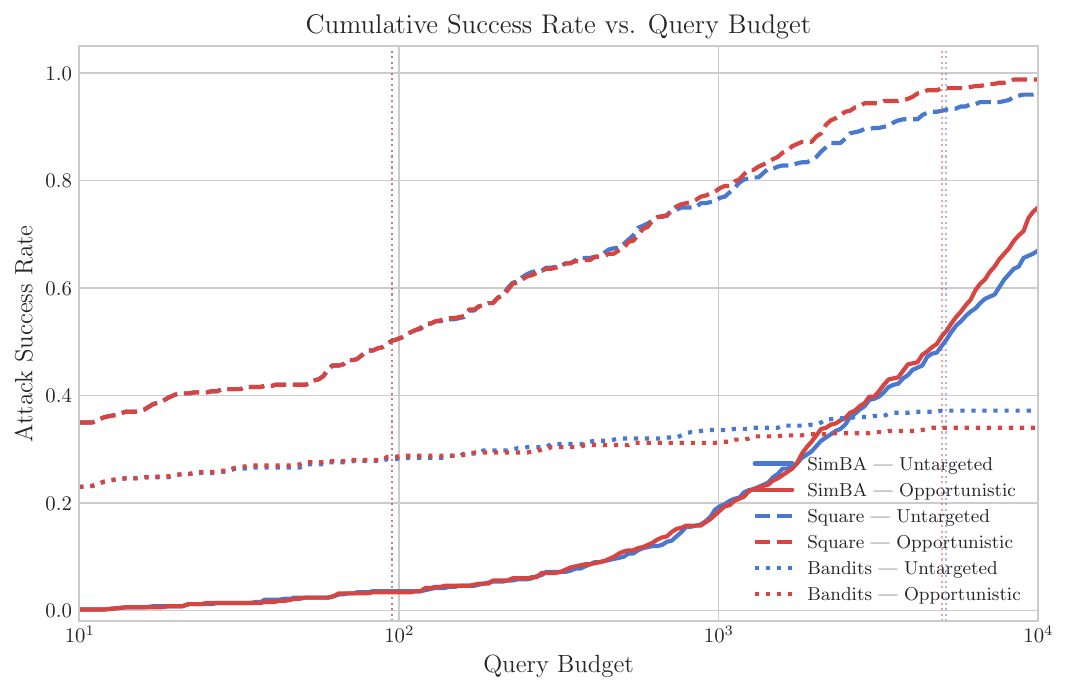}
\caption{Success rate vs.\ query budget (100-image benchmark, all 5 standard models pooled). SimBA and Square Attack (CE) show clear OTS benefit. Bandits (dotted) shows no separation, confirming that gradient estimation already provides directionality.}
\label{fig:cdf-pooled}
\end{figure}

\subsection{Iteration Efficiency}\label{sec:iteration-efficiency}

All Wilcoxon tests are two-sided signed-rank tests with Bonferroni correction across models.

\textbf{Censored mean} (failed runs assigned their respective budget ceiling). SimBA: OTS reduces the censored mean by 4.8\% (5,426 $\to$ 5,164). Square Attack (CE): 29.5\% reduction (1,076 $\to$ 759), largely driven by converting failures to successes rather than faster convergence on already-successful runs. Bandits: slight increase of 3.7\% (3,254 $\to$ 3,374). SimBA and Square Attack show highly significant pooled Wilcoxon tests ($p < 10^{-9}$, Bonferroni-corrected). For Square Attack, per-model significance concentrates on ResNet-50 and ViT-B/16; easy models show floor effects ($<$100 iterations regardless of mode). For Bandits, no comparison reaches significance.

\begin{figure}[ht]
\centering
\includegraphics[width=0.85\textwidth]{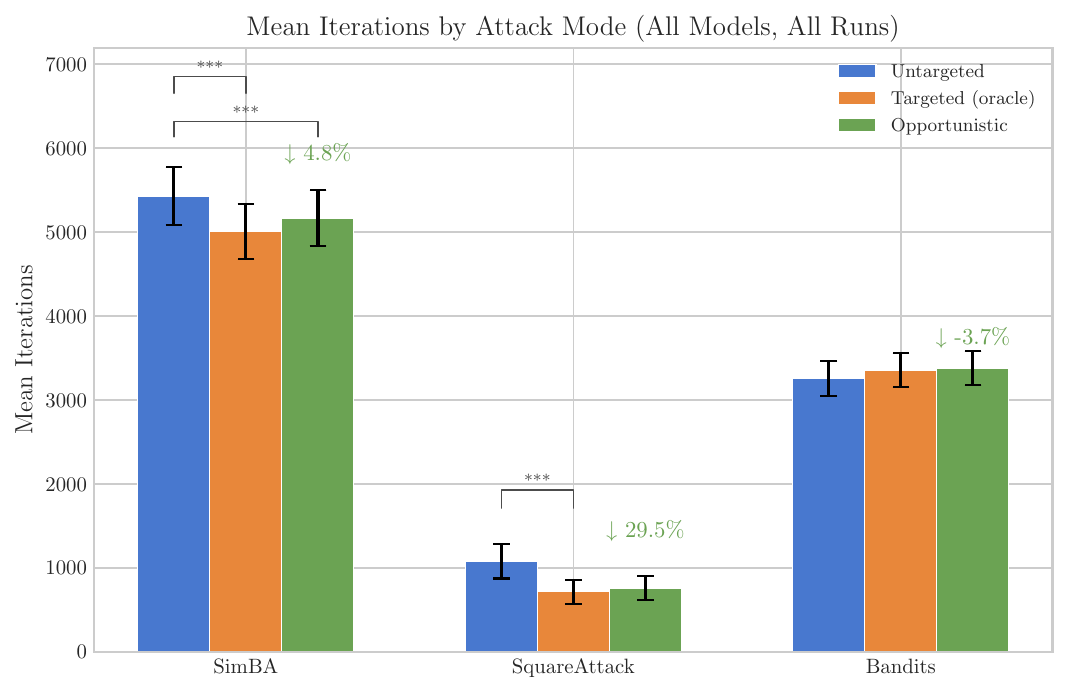}
\caption{Censored mean iterations by attack mode (4,500 runs; failed runs assigned 10K ceiling). Error bars show 95\% bootstrap CI. Significance brackets show Bonferroni-corrected Wilcoxon signed-rank tests (pooled across models).}
\label{fig:headline-bars}
\end{figure}

\subsection{Difficulty-Dependent Benefit}\label{sec:depth-scaling}

The largest gains appear on ResNet-50 and ViT-B/16, the two hardest models. \textit{Attack difficulty}, not model depth per se, is the driver: ViT-B/16 (12 transformer blocks, high OTS benefit) and VGG-16 (16 convolutional layers, negligible OTS benefit) show that depth alone does not predict the effect. The difficulty--savings scatter plot (Appendix~\ref{app:secondary-figures}) shows a weak but statistically significant positive correlation ($r = 0.19$, $p < 0.001$, $3.6\%$ variance explained). Per-model breakdowns are in Appendix~\ref{app:secondary-figures}.

\subsection{Lock-in Dynamics}\label{sec:lockin}

Lock-in occurs early (SimBA median: 12.5 iterations; Square Attack median: 141), and exact lock-match with the oracle class is not required: OTS closes nearly the entire untargeted-to-oracle gap because it only needs \textit{a viable} class, not the oracle class. Detailed lock-in traces and lock-match analysis are in Appendix~\ref{app:secondary-figures}.

\section{OTS as a Margin Surrogate}\label{sec:margin-surrogate}

\subsection{Equivalence Under Stable Rankings}\label{sec:equivalence}

Once OTS locks onto target class $t$, the attack minimizes $-P(t|x')$. If $t = \arg\max_{k \neq y} P(k|x')$ at lock-in, then maximizing $P(t|x')$ is equivalent to maximizing $\max_{k \neq y} P(k|x')$, which is exactly the margin loss's competitor term. The margin loss decomposes as:
\begin{equation}
\mathcal{L}_{\text{margin}} = \underbrace{f_y(x')}_{\text{push down true class}} - \underbrace{f_t(x')}_{\text{push up competitor}}
\end{equation}
where $t$ is dynamically reselected each iteration. OTS approximates this by fixing $t$ after the exploration phase. The approximation is tight when the locked class remains the strongest competitor throughout the attack. Note that the equivalence between maximizing $P(t|x')$ and maximizing $f_t(x')$ is approximate: softmax is monotonic in each logit with others fixed, but perturbation steps change all logits simultaneously. In practice, the locked class's logit dominates the update direction after lock-in, and the empirical results confirm the approximation is effective.

\subsection{Empirical Confirmation: CE Loss Ablation}\label{sec:loss-ablation}

\begin{table}[H]
\centering
\caption{Loss function ablation (Square Attack, ResNet-50, 15K budget).}
\label{tab:loss-ablation}
\small
\begin{tabular}{@{}llll@{}}
\toprule
Configuration & Success Rate & Mean Iter. & Median Iter. \\
\midrule
CE untargeted & 85.0\% & 2,601 & 335 \\
CE + OTS & 98.0\% & 1,791 & 756 \\
CE oracle targeted & 99.0\% & 1,865 & 640 \\
Margin untargeted & 98.7\% & 1,453 & 366 \\
Margin + OTS & 98.7\% & 1,583 & 396 \\
\bottomrule
\end{tabular}
\end{table}

Table~\ref{tab:loss-ablation} confirms OTS's role as a margin surrogate. Square Attack with margin loss shows \textbf{no benefit} from OTS (Wilcoxon $p = 0.08$, $N = 74$): margin loss already performs dynamic target tracking at every iteration. Switching to CE loss degrades the untargeted attack from 98.7\% to 85.0\% success; OTS restores CE to near-margin performance (98.0\% success). Note that OTS's higher median iteration count (756 vs.\ 335) reflects a survivor bias: OTS converts 13 additional failures into successes, and these hard images converge slowly, raising the conditional median among successes. The Bandits negative result (Section~\ref{sec:success-rates}) provides independent confirmation. This establishes a general principle: \textbf{OTS helps attacks that lack implicit target tracking, and is unnecessary for attacks that already have it}. Crucially, OTS cannot simply be replaced by switching the loss to margin: margin loss requires the actual logit values $f_y(x')$ and $\max_{k \neq y} f_k(x')$ at every iteration, whereas OTS only needs a single $\arg\max_{k \neq y}$ at lock-in time. This makes OTS applicable to any attack that exposes a class ranking, regardless of whether it internally operates on logits, probabilities, or labels.

\subsection{Perturbation Alignment}\label{sec:theta-convergence}

Inspired by the angular convergence framework of \citet{gesny2024gradient}, we track the cosine similarity between each attack's perturbation $\delta(i) = x'(i) - x$ and the oracle direction $\delta_{\text{oracle}}$ (the perturbation produced by a targeted attack toward the oracle class). We use a 500-iteration horizon to focus on the alignment dynamics around the switch point; alignment stabilizes well before this budget.

\begin{figure}[ht]
\centering
\includegraphics[width=0.6\textwidth]{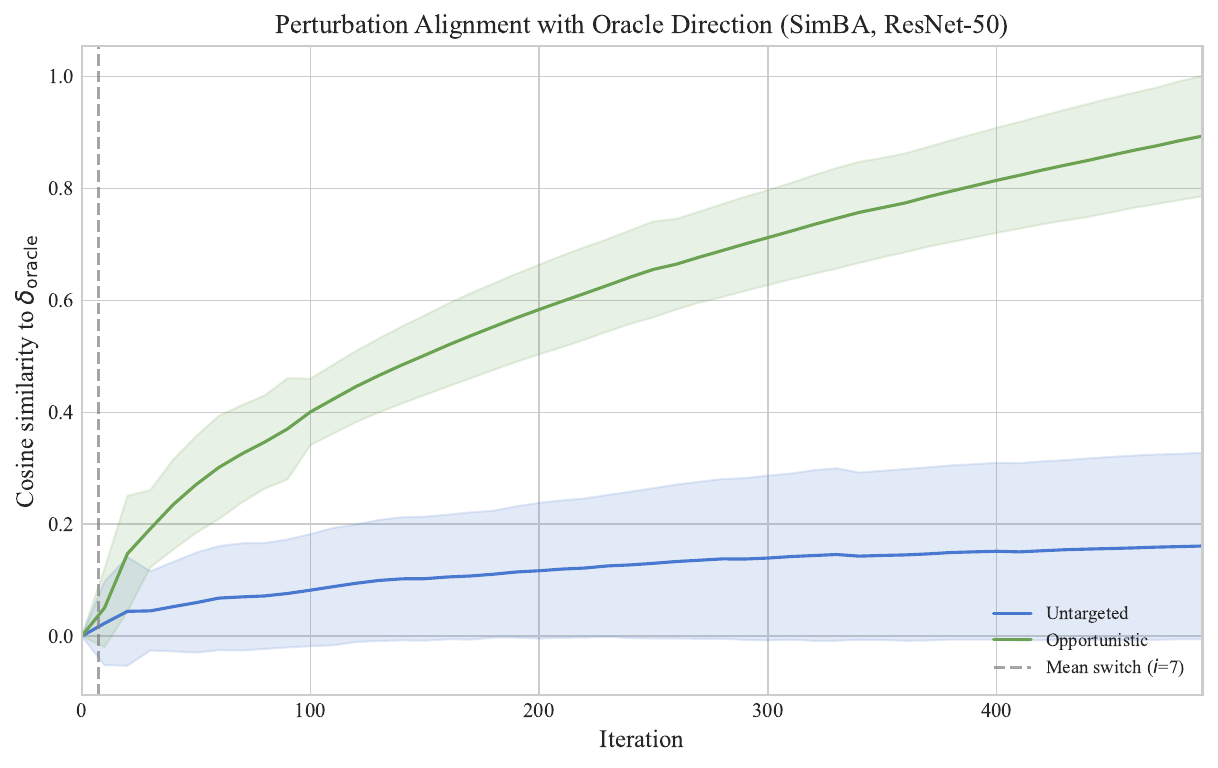}
\caption{Perturbation alignment with oracle direction (SimBA, ResNet-50, 100 images, 500-iteration budget). Shaded regions: $\pm 1$ std.\ dev. Vertical dashed line: mean switch iteration.}
\label{fig:theta}
\end{figure}

Untargeted perturbations reach a terminal cosine similarity of only $0.174 \pm 0.189$ ($\theta \approx 80^\circ$). Opportunistic perturbations, after switching at mean iteration 7.3, rapidly align with $\delta_{\text{oracle}}$, reaching $0.865 \pm 0.192$ ($\theta \approx 30^\circ$). This alignment gap of $0.692$ demonstrates that OTS actively redirects the perturbation toward the oracle basin, not just selecting the correct target class.

\section{Robust Models}\label{sec:robust}

We extend the evaluation to adversarially-trained ImageNet classifiers from RobustBench \citep{salman2020adversarially,croce2021robustbench}: Salman2020Do\_R18 and Salman2020Do\_R50 (50 images, $S=10$). Success rates are flat across modes: SimBA achieves 22--28\% and Square Attack 56--70\%, with no statistically significant difference between untargeted, oracle, and OTS (all Bonferroni-corrected Wilcoxon $p > 0.09$). Sweeping $S \in \{10, \ldots, 20\}$ does not change the picture (Appendix~\ref{app:ablation-details}).

\textbf{Interpretation.} The explanation lies in the difficulty distribution, not landscape smoothing. On standard ResNet-50, images span a range of difficulties including a ``medium'' zone (100--5,000 iterations) where directional commitment pays off. On robust ResNet-50, the distribution is \textbf{bimodal}: images either succeed in $<$100 iterations or fail at the budget ceiling, with only 9--10 images in the medium zone. The failure set is nearly identical across all three modes (22/21/21 failures). Targeting cannot rescue genuinely infeasible images; adversarial training compresses the difficulty spectrum into two extremes.

\section{Discussion and Conclusion}\label{sec:conclusion}

\subsection{Ablations}\label{sec:ablations}
Sweeping the stability threshold $S \in \{2, \ldots, 15\}$ on ResNet-50 (100 images, 15K budget) shows at most 2.1 pp variation in success rate (Appendix~\ref{app:ablation-details}). A stronger test removes the stability mechanism entirely: a \textbf{fixed-iteration switch} at iteration $T$ with no stability check yields indistinguishable results for all $T \in \{5, \ldots, 500\}$. We recommend the fixed-iteration switch as the default.

Two further ablations isolate what drives OTS's benefit. First, \textbf{targeting a random non-true class} is catastrophic: success drops to 14\% for SimBA and 53\% for Square Attack (CE), far below untargeted baselines (43\% and 81\%). Target \emph{selection} matters, not just directional commitment. Second, \textbf{targeting the clean-image argmax} ($T = 0$, no exploration) nearly matches OTS: 79\% vs.\ 80\% for SimBA and 96\% vs.\ 96\% for Square Attack (CE). However, Square Attack with a short exploration ($T = 5$) reaches 99\%, a 3~pp gain over $T = 0$. Since the exploration phase costs less than 1\% of a typical budget, it provides cheap insurance against edge cases where early perturbation steps reveal a better target than the clean-image ranking.

A \textbf{target-class overlap analysis} (Figure~\ref{fig:target-overlap}) clarifies why: SimBA locks onto the clean-image top-1 in 100\% of runs (small coordinate perturbations preserve the ranking), while Square Attack shows lower overlap (64\% at top-1) because random patches are more disruptive. This explains why exploration helps Square Attack but not SimBA.

\subsection*{Limitations and Future Work}\label{sec:limitations}
Two main limitations should be noted. First, our evaluation on robust models is restricted to two architectures trained under an $L_\infty$ adversarial setting, and the observed behavior may not generalize to other defense mechanisms such as randomized smoothing or input transformations. Second, all experiments are conducted on ImageNet ($K = 1{,}000$), a regime where class drift is particularly pronounced due to the large number of alternative classes; as a result, the benefits of OTS may be reduced in settings with fewer classes where the probability mass has less room to scatter. Natural extensions include testing on CIFAR-10 ($K=10$) and under $L_2$ budgets to probe whether drift dynamics differ, as well as an adaptive re-locking variant that releases the lock when the locked class drops in ranking.
\bibliography{references}

\clearpage
\appendix
\section*{Appendix}

\section{Secondary Figures}\label{app:secondary-figures}

\begin{figure}[htbp]
\centering
\includegraphics[width=0.85\textwidth]{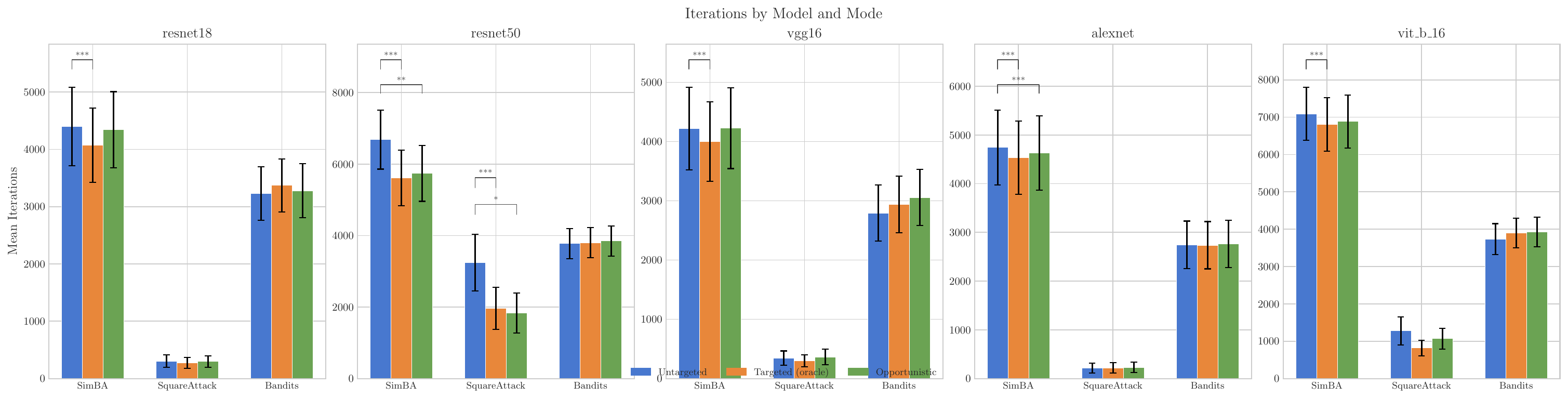}
\caption{Iterations by model and mode for SimBA, Square Attack (CE), and Bandits. OTS's benefit is largest on harder models (ResNet-50, ViT-B/16) and absent for Bandits.}
\label{fig:per-model}
\end{figure}

\begin{figure}[htbp]
\centering
\includegraphics[width=0.85\textwidth]{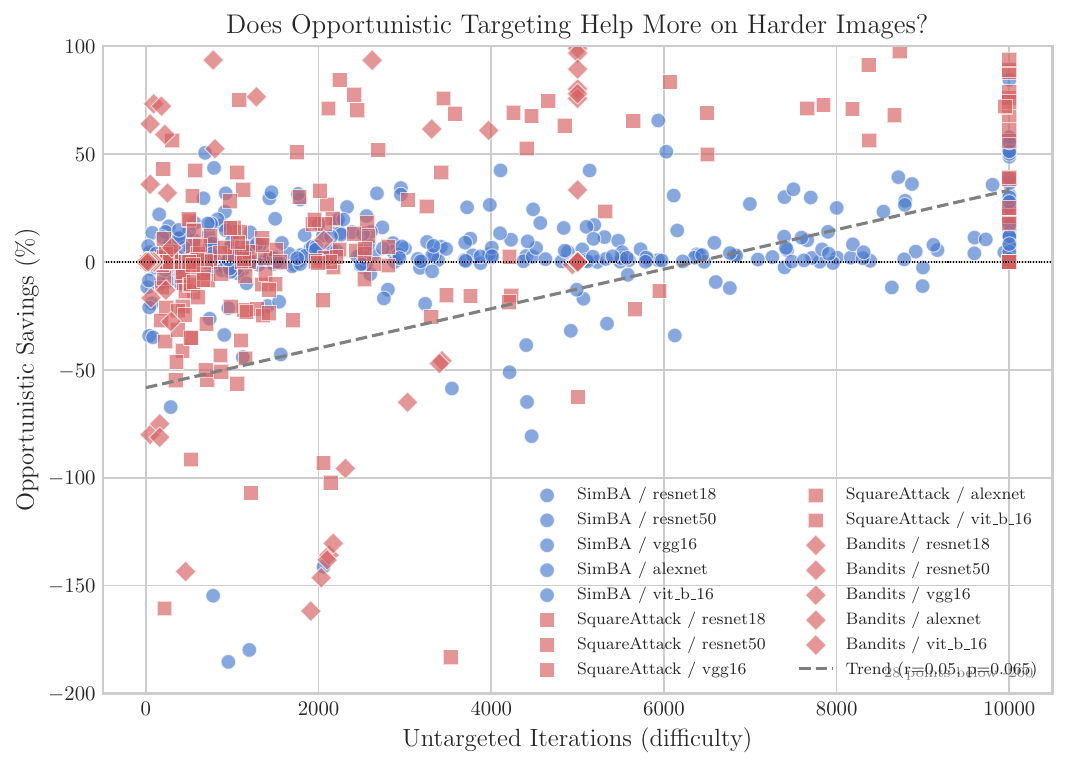}
\caption{Opportunistic savings vs.\ attack difficulty. Each point is one (model, method, image) run. Dashed line: linear trend ($r = 0.19$, $p < 0.001$).}
\label{fig:difficulty-savings}
\end{figure}

\begin{figure}[htbp]
\centering
\includegraphics[width=0.85\textwidth]{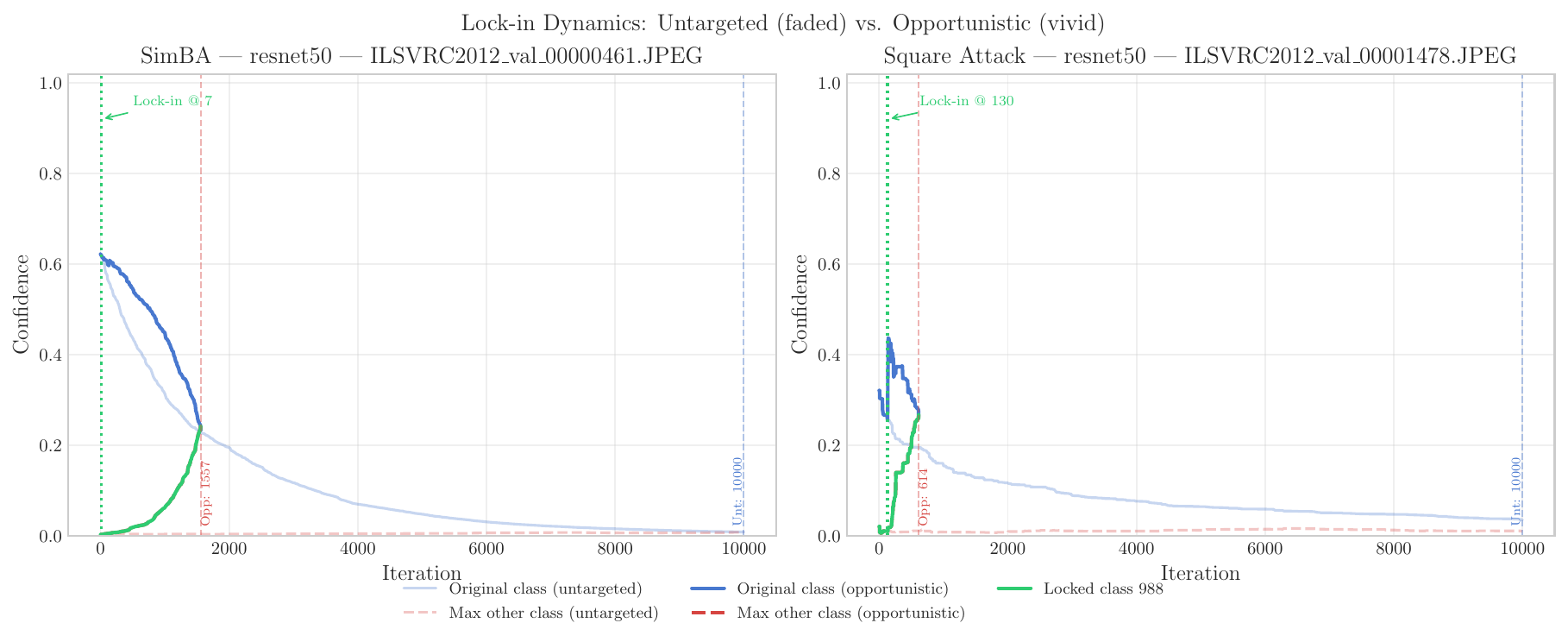}
\caption{Lock-in dynamics. Faded curves: untargeted; vivid: opportunistic. Vertical dotted lines mark lock-in and convergence iterations. Cases selected as highest-savings image per method from the 50-image benchmark on ResNet-50.}
\label{fig:lockin}
\end{figure}

\begin{figure}[htbp]
\centering
\includegraphics[width=0.85\textwidth]{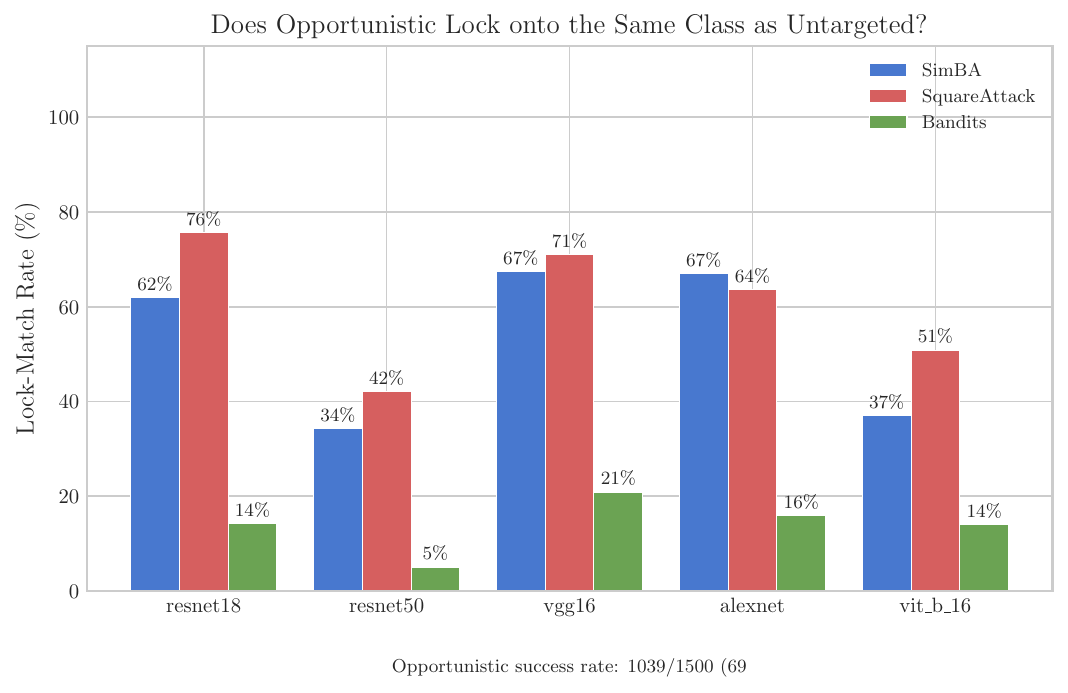}
\caption{Lock-match rate by model and method. SimBA and Square Attack show moderate rates (35--75\%). Bandits shows very low rates (5--21\%).}
\label{fig:lock-match}
\end{figure}

\clearpage
\section{Ablation Details}\label{app:ablation-details}

\begin{table}[htbp]
\centering
\caption{Stability threshold sweep (ResNet-50, 100 images, 15K budget). Bold: selected $S$.}
\label{tab:s-sweep}
\small
\begin{tabular}{@{}llll@{\hspace{1.5em}}llll@{}}
\toprule
\multicolumn{4}{c}{SimBA} & \multicolumn{4}{c}{Square Attack (CE)} \\
\cmidrule(r){1-4} \cmidrule(l{-0.6em}){5-8}
$S$ & Succ. & Mean & Median & $S$ & Succ. & Mean & Median \\
\midrule
2 & 84.0\% & 4,860 & 4,108 & 2 & 97.1\% & 1,917 & 779 \\
3 & 84.2\% & 4,814 & 3,905 & 3 & 97.1\% & 1,929 & 774 \\
5 & 85.0\% & 4,952 & 4,286 & 5 & 97.1\% & 2,004 & 754 \\
8 & 85.1\% & 4,915 & 4,144 & \textbf{8} & \textbf{98.1\%} & \textbf{1,780} & \textbf{753} \\
\textbf{10} & \textbf{85.1\%} & \textbf{4,889} & \textbf{4,075} & 10 & 96.1\% & 1,719 & 582 \\
12 & 84.0\% & 4,832 & 4,143 & 12 & 97.0\% & 1,850 & 633 \\
15 & 85.0\% & 4,956 & 4,366 & 15 & 96.0\% & 1,910 & 652 \\
\bottomrule
\end{tabular}
\end{table}

\begin{figure}[htbp]
\centering
\includegraphics[width=0.85\textwidth]{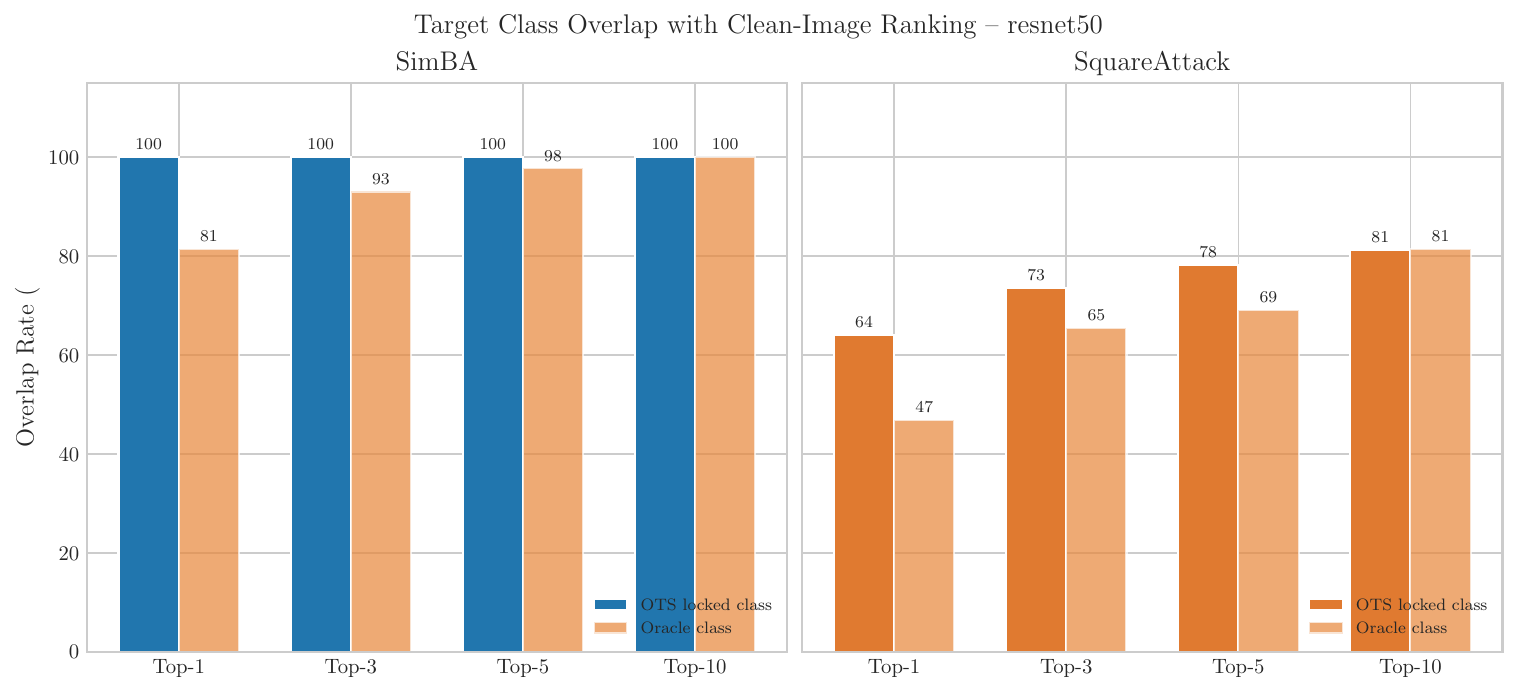}
\caption{Overlap between OTS's locked class (or the trajectory oracle class) and the clean-image top-$K$ non-true classes (ResNet-50). SimBA always locks onto the clean-image top-1; Square Attack's random-patch perturbations disrupt the ranking more, explaining why a short exploration phase helps for Square Attack but not SimBA.}
\label{fig:target-overlap}
\end{figure}

\begin{figure}[htbp]
\centering
\includegraphics[width=0.7\textwidth]{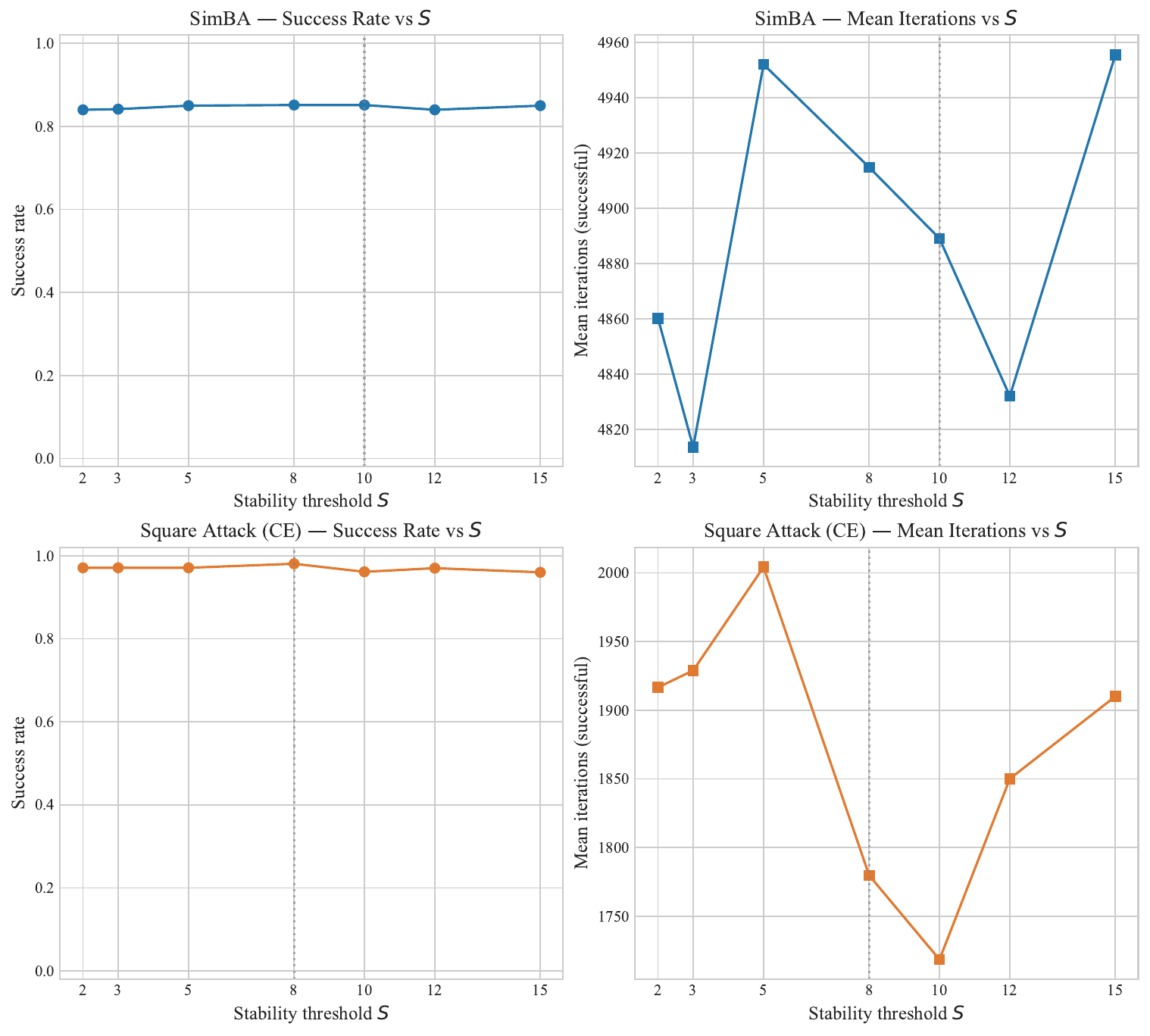}
\caption{Stability threshold ablation. Top: SimBA, bottom: Square Attack (CE). Left: success rate vs.\ $S$; right: mean iterations vs.\ $S$. Dotted lines mark selected $S$.}
\label{fig:s-ablation}
\end{figure}

\begin{figure}[htbp]
\centering
\includegraphics[width=0.85\textwidth]{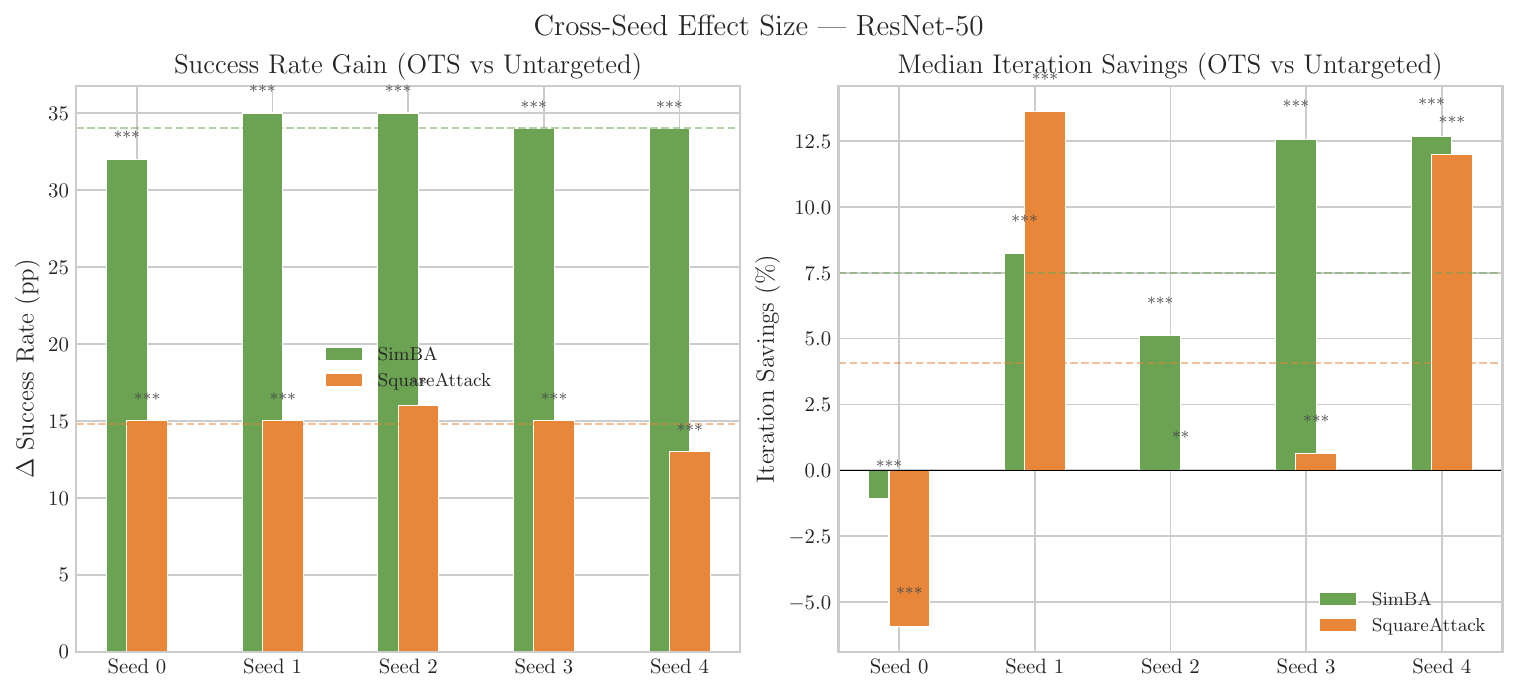}
\caption{Multi-seed validation (ResNet-50, 15K budget, 5 seeds $\times$ 100 images). Left: success rate gain (OTS vs.\ untargeted) per seed. Right: median iteration savings per seed. Stars denote Bonferroni-corrected Wilcoxon significance. All 10 tests reach $p < 0.002$, confirming seed independence.}
\label{fig:multiseed}
\end{figure}

\end{document}